# TRACE: Reconstruction-Based Anomaly Detection in Ensemble and Time-Dependent Simulations

Hamid Gadirov, Martijn Westra, Steffen Frey

**Abstract.** Detecting anomalous behavior in high-dimensional, time-dependent simulation data is an important yet challenging task in scientific computing and visualization. In this work, we investigate reconstruction-based anomaly detection for ensemble data generated from parameterized Kármán vortex street simulations using convolutional autoencoders. We conduct a systematic comparison between two architectural variants: a two-dimensional convolutional autoencoder that operates on individual time steps, and a three-dimensional convolutional autoencoder that processes short temporal stacks of consecutive simulation frames. Our results show that the 2D autoencoder is effective at identifying localized spatial irregularities within individual images, capturing anomalies that manifest as deviations in instantaneous flow structure. In contrast, the 3D autoencoder leverages spatio-temporal context and is able to associate patterns across time, enabling the detection of anomalies related to dynamic behavior and motion characteristics in the simulation. This temporal awareness allows the 3D model to identify anomalous evolution patterns that are not apparent when analyzing frames independently. We further evaluate reconstruction accuracy on time-dependent volumetric data and observe that reconstruction errors are strongly influenced by the spatial distribution of mass within the volume. In particular, configurations with highly concentrated mass in localized regions consistently yield higher reconstruction errors than cases with more spatially dispersed mass distributions. This finding highlights the sensitivity of reconstruction-based methods to localized extremes and provides insight into the types of physical structures that are likely to be flagged as anomalous. Overall, this study demonstrates the complementary strengths of 2D and 3D convolutional autoencoders for anomaly detection in ensemble and time-dependent simulation data, and underscores the importance of incorporating temporal context when analyzing dynamic flow phenomena.

## 1. Introduction

In recent years, the rapid growth of high-dimensional scientific simulation data has created new opportunities for discovering complex patterns, while simultaneously introducing significant challenges for analysis and interpretation. In particular, ensemble simulations—collections of simulations generated by varying initial conditions or model parameters—are widely used in domains such as climate science, fluid dynamics, and uncertainty quantification. These ensembles often contain subtle, rare, or unexpected behaviors that are difficult to detect using traditional statistical or visualization techniques.

One promising approach for addressing these challenges is anomaly detection. Anomalies in ensemble data may correspond to rare physical phenomena, numerical instabilities, or specific regions of parameter space that produce qualitatively different behavior. Identifying such anomalies can provide valuable insights into the underlying simulation models and guide further investigation of parameterizations or model assumptions. In addition to ensemble-based analysis, anomaly detection is also highly relevant for time-dependent simulations, where unexpected transitions or events may occur at specific time steps within a single simulation run.

To detect anomalies in both ensemble and time-varying simulation data, we explore the use of convolutional autoencoders (CAEs). Autoencoders are neural networks composed of two main components: an encoder, which maps high-dimensional input data to a lower-dimensional latent representation, and a decoder, which reconstructs the original data from this compressed representation. In the case of convolutional autoencoders, convolutional layers are used to effectively capture spatial structure in image- or volume-based scientific data.

During training, autoencoders learn to reconstruct input samples as accurately as possible by minimizing a reconstruction loss. As a result, the latent space captures the most salient features that are common across the training data. When presented with previously unseen or rare patterns, the model often struggles to reconstruct them accurately. This reconstruction difficulty can be quantified using reconstruction error metrics, which form the basis for anomaly detection. Samples with unusually high reconstruction errors are flagged as potential anomalies, as their underlying patterns were absent or underrepresented during training.

In the context of ensemble data, detected anomalies can be directly linked back to the individual simulations from which they originate. This enables targeted analysis of the corresponding parameter settings, potentially revealing relationships between simulation parameters and anomalous behavior. While deeper investigation into the physical or numerical causes of such anomalies lies beyond the scope of this work, our framework provides a systematic way to identify candidate cases for further study.

Beyond ensemble-level analysis, we also apply this approach to time-dependent simulations. By evaluating reconstruction errors at successive time steps, we can identify moments in time where the autoencoder fails to accurately reconstruct the simulated fields. Such time-localized anomalies may correspond to transitions, instabilities, or other events of interest within the simulation, offering a data-driven method to highlight important temporal regions without prior domain-specific thresholds.

Overall, this work demonstrates how autoencoder-based feature learning and reconstruction analysis can serve as a unified framework for anomaly detection in both ensemble and time-varying scientific simulation data. By leveraging the representational power of convolutional autoencoders, we aim to provide an effective and scalable approach for identifying rare and informative patterns in complex, high-dimensional datasets.

2. **Related Work**

This section reviews prior work relevant to autoencoder-based representation learning and its application to anomaly detection and ensemble analysis in scientific data. We focus on three closely related areas: autoencoder-based feature learning, deep learning for scientific ensemble data, and reconstruction-based anomaly detection in spatio-temporal simulations.

**2.1 Autoencoder-Based Feature Learning**

Autoencoders are a widely used class of neural networks for learning compact representations of high-dimensional data. By training an encoder–decoder architecture to reconstruct its input, an autoencoder learns a latent representation that preserves the most salient structures of the data distribution. Advances in deep learning, including the introduction of rectified linear units (ReLU) [21] and regularization techniques such as dropout [4], enabled stable training of deeper autoencoder architectures and improved generalization.

In the context of scientific data analysis and visualization, autoencoders have proven effective for dimensionality reduction of spatially structured datasets. Gadirov et al. systematically evaluated autoencoder architectures for expressive dimensionality reduction of spatial ensemble data, demonstrating that convolutional autoencoders can produce latent spaces that preserve physically meaningful variability across ensemble members [1]. This work established design principles for encoder–decoder architectures tailored to scientific ensembles and highlighted their advantages over classical linear methods. These findings were further explored in subsequent work on autoencoder-based feature extraction for ensemble visualization [7, 24].

Autoencoders have also been applied to volumetric and time-dependent data. ENTIRE introduces an autoencoder-based framework that learns structure-aware features from volumetric fields and combines them with rendering parameters to predict volume rendering time [9]. While ENTIRE focuses on performance prediction rather than data analysis, it demonstrates that autoencoders can extract informative representations from complex spatio-temporal simulation data, reinforcing their suitability for scientific workflows.

**2.2 Deep Learning for Scientific Ensemble Data**

Scientific ensemble datasets arise from simulations or measurements performed under varying parameters or initial conditions. Analyzing such ensembles requires methods that can capture both spatial structure and variability across ensemble members. Autoencoder-based approaches are particularly well suited for this task because they learn compact latent representations that encode dominant patterns while remaining flexible across simulation configurations.

Beyond static ensemble representations, recent work has focused on learning temporal structure within ensembles. FLINT introduces a learning-based approach for flow estimation and temporal interpolation in 2D+time and 3D+time ensemble data, reconstructing velocity fields and scalar quantities without strong domain assumptions [6]. HyperFLINT extends this approach by using a hypernetwork conditioned on simulation parameters, improving generalization across ensemble configurations and enabling parameter-aware temporal interpolation [5]. These methods show that neural networks can successfully learn spatial and temporal coherence in ensemble simulations and generalize across parameter spaces.

A broader overview of machine learning techniques for scientific visualization and ensemble analysis is provided in recent dissertation work, which discusses autoencoder-based dimensionality reduction, learning-based flow estimation, and hypernetwork approaches as complementary tools for analyzing complex spatio-temporal ensembles [8]. Together, these studies establish learned latent representations as a central component of modern ensemble data analysis pipelines.

**2.3 Reconstruction-Based Anomaly Detection in Spatio-Temporal Data**

Reconstruction-based anomaly detection is a well-established paradigm in machine learning. In this setting, a model is trained to reconstruct typical or "normal" data patterns, and samples that deviate from these patterns are identified via elevated reconstruction errors. Autoencoders are particularly well suited for this approach, as their latent spaces emphasize common structures while failing to generalize to rare or unseen configurations.

Such methods have been successfully applied to multivariate time series and high-dimensional data, including approaches that combine autoencoders with contrastive objectives for improved separation of normal and abnormal patterns [22]. In scientific simulation data, however, anomaly detection remains comparatively underexplored, especially for high-dimensional spatial fields and ensemble datasets. Unlike classical anomaly detection problems, anomalies in scientific simulations may correspond to physically meaningful but rare phenomena rather than noise or faults.

Our work builds on the observation that autoencoders trained on ensemble or time-dependent simulation data implicitly encode dominant structural patterns of the data distribution. By analyzing reconstruction errors across ensemble members or time steps, anomalies can be identified and linked back to specific simulations or temporal regions. This complements prior work on parameter-space exploration and region transitions in simulation data [20], while providing a fully data-driven mechanism for highlighting atypical behavior.

## 2.4 Implementation and Computational Context

Modern deep learning frameworks such as PyTorch provide efficient automatic differentiation and scalable training for autoencoder architectures [23]. Learned latent representations are often further analyzed using nonlinear dimensionality reduction techniques such as UMAP [25] to support visual exploration of ensemble structure [22]. Training and evaluation of such models typically require substantial computational resources, which are provided by high-performance computing infrastructures such as the Peregrine HPC cluster at the University of Groningen [2].

To ensure reproducibility and portability of machine learning pipelines, containerization technologies such as Docker are commonly used in scientific computing environments [18, 19]. These practices align with broader trends toward scalable, reproducible, and system-integrated learning-based analysis in scientific visualization workflows.

3. Methods and Experiments

## 3.1 2D Convolutional Autoencoder

**Data and Pre-processing.** To study anomaly detection in ensemble simulation data, we use a parameterized Kármán vortex street dataset introduced by Fernandes et al. This dataset consists of numerical simulations that exhibit a characteristic pattern of alternating vortices forming downstream of an obstacle. Each simulation represents a different parameterization of the same underlying physical system. An example of a simulation state at a given time step is shown in Figure 1.

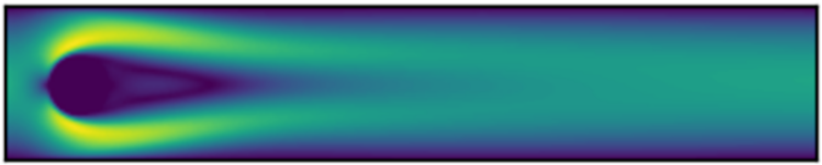

Figure 1: Kármán vortex street simulation volume

The dataset contains 300 individual simulations, each consisting of 54 time steps. All simulations start from an empty domain, and the vortex street pattern gradually develops over the initial frames. As the early time steps primarily capture the formation phase rather than the fully developed flow, we discard the first 18 time steps of each simulation. This results in 36 usable time steps per simulation that contain stable and informative flow patterns.

From the full ensemble, we select a subset of 90 simulations, yielding a total of 3,240 images (90 simulations × 36 time steps). These images are randomly sampled across simulations and time steps. We empirically found that random sampling leads to more accurate and stable reconstructions compared to selecting consecutive simulations, as it exposes the model to a broader range of spatial variations during training. The selected data are split into training, validation, and test sets in a 70 % / 15 % / 15 % ratio.

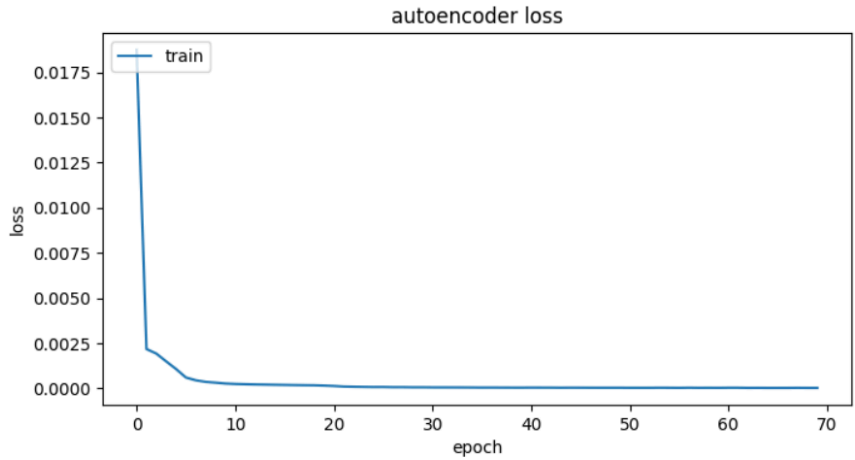

Figure 2: Validation loss in training the model

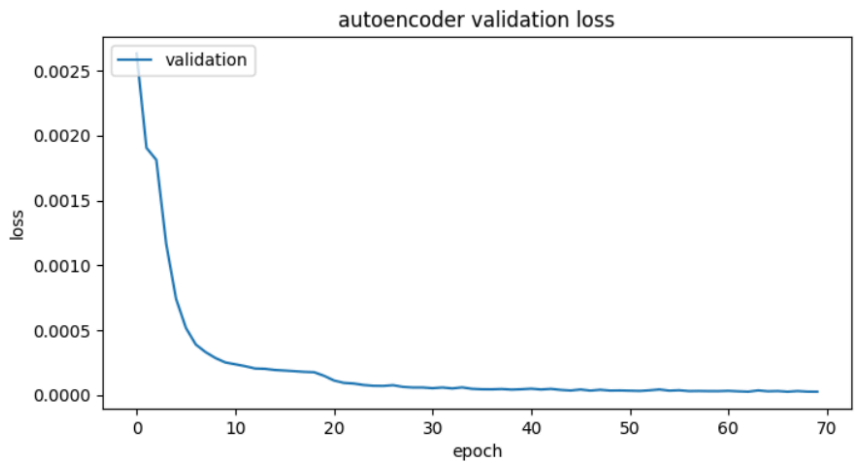

Figure 3: Validation loss in training the model

We take a small selection of images from the testing set and compare reconstructions created by the model to the originals. These are shown in Figure 4.

All images are normalized to the range [0, 1]. Rather than directly scaling the original byte values, we compute the global minimum and maximum over the dataset and normalize each value by subtracting the minimum and dividing by the corresponding range. This approach is motivated by the observation that the data do not fully occupy the original byte range: most values are concentrated below mid-range intensities. Simple division by 255 therefore results in poor utilization of the normalized space and inferior reconstruction quality. The pre-processed data are stored in serialized format to facilitate efficient loading during training and evaluation.

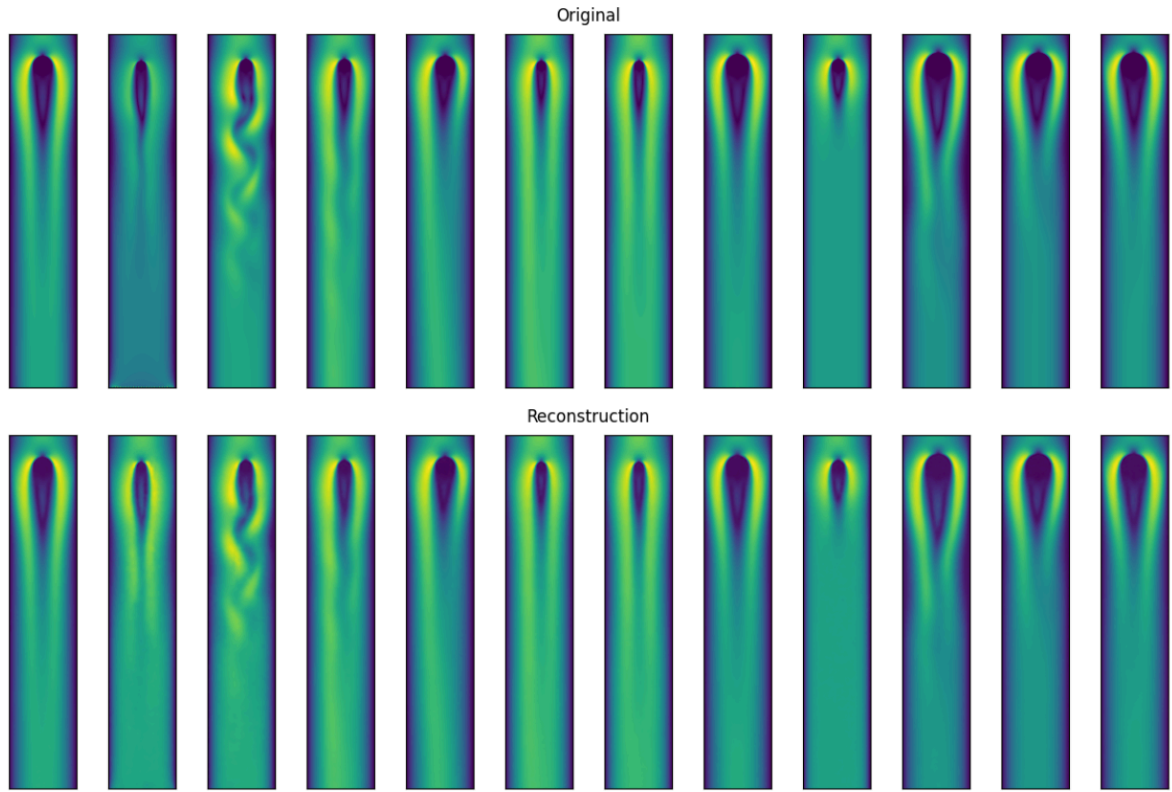

Figure 4: Reconstructions versus originals at 70 epochs

**Model Configuration.** The 2D convolutional autoencoder is implemented using PyTorch. The encoder consists of four convolutional layers, each employing 64 filters with a kernel size of 3 × 3 and a stride of 2, progressively reducing the spatial resolution while increasing representational abstraction. The output of the final convolutional layer is flattened and passed through two fully connected layers with 256 and 128 units, respectively, forming the latent representation.

The decoder mirrors the encoder structure. A dense layer first expands the latent vector back to the dimensionality of the final convolutional feature map, after which the data are reshaped to a two-dimensional tensor. A sequence of transposed convolutional layers is then applied to reconstruct the original spatial resolution. Due to the downsampling and upsampling operations, the reconstructed output is slightly larger than the input; we therefore apply a cropping operation to restore the original dimensions. A detailed description of the model architecture is provided in Appendix A, with the full Keras model summary included in Appendix B.

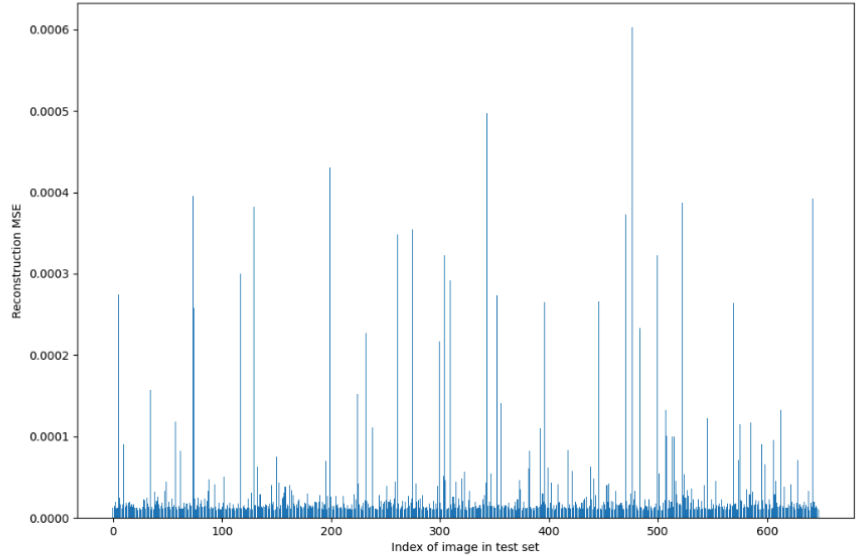

Figure 5: Reconstruction MSE for images in test set

**Training Procedure.** The model is trained using the training subset while monitoring reconstruction loss on the validation set. Mean squared error (MSE) is used as the reconstruction objective. Training is performed until convergence of the validation loss, which typically occurs after approximately 70 epochs. The loss curve is shown in Figure 2. The validation loss curve is shown in Figure 3. The early stopping criterion helps prevent overfitting while ensuring sufficient learning capacity to capture dominant flow structures.

Testing and Anomaly Detection. To assess reconstruction quality on unseen data, we compute the reconstruction MSE for each image in the test set. The resulting error values are visualized as a bar plot (Figure 5), which reveals a small number of samples with significantly higher reconstruction errors than the majority of the data.

We define an anomaly threshold of $3 \times 10^{-4}$ to isolate these extreme cases. Images whose reconstruction error exceeds this threshold are classified as anomalous and reconstructed using the trained autoencoder. Representative examples are shown in Figure 6.

Inspection of the anomalous samples reveals that several of these images exhibit localized or irregular turbulent patterns. In some cases, the original data contain low-magnitude values—visually indicated by blue or purple regions—yet display spatial structures that resemble higher-pressure or higher-energy flow configurations. Since such patterns are uncommon in the training data, the autoencoder tends to reconstruct them with elevated values, leading to higher reconstruction errors. Other anomalous samples show more pronounced turbulence or structural complexity, which the autoencoder struggles to reproduce faithfully. These observations confirm that the reconstruction-based approach effectively highlights spatial patterns that deviate from the dominant behavior learned during training.

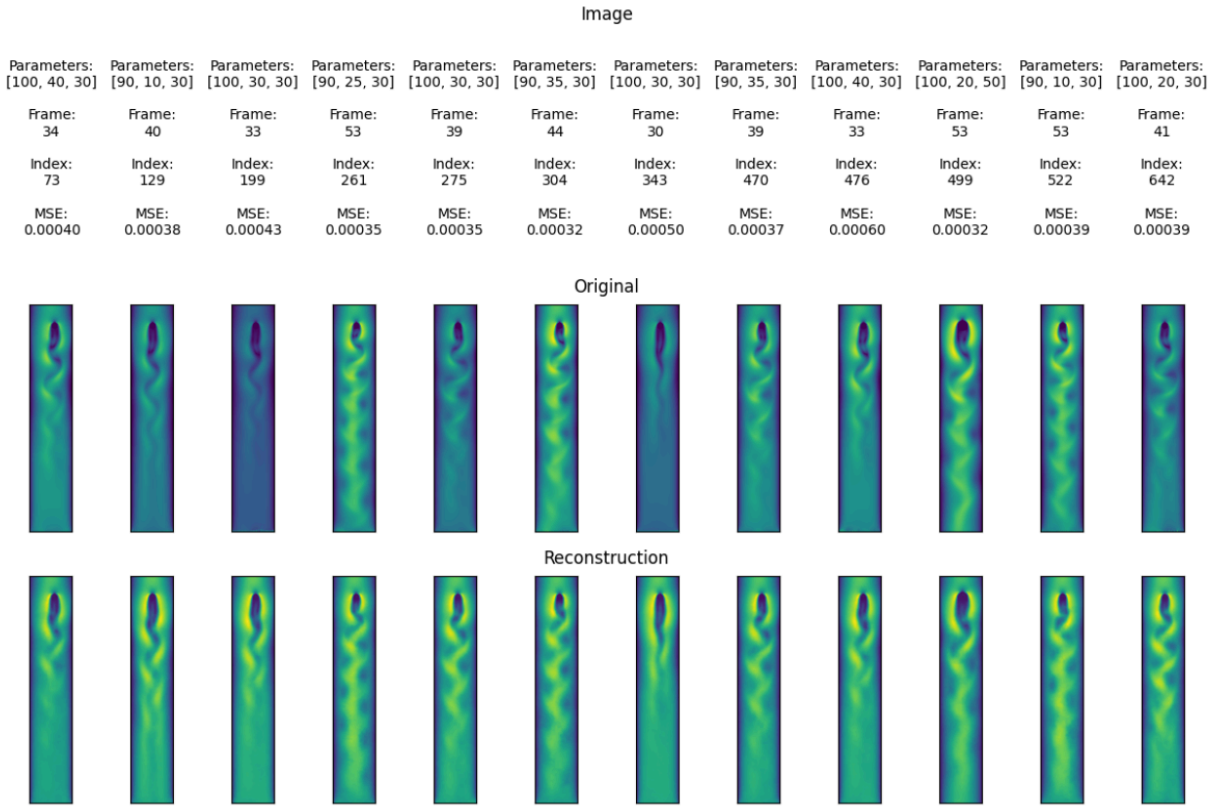

Figure 6: Reconstruction for images with high MSE in the test set

### 3.2 3D Convolutional Autoencoder for the Kármán Vortex Street Ensemble

**Data Pre-processing.** To incorporate temporal context into the anomaly detection process, we extend the 2D convolutional autoencoder to a 3D formulation. Instead of operating on individual simulation frames, the 3D autoencoder processes short temporal volumes constructed by stacking three consecutive time steps along a third dimension. This representation allows the model to capture local temporal coherence and motion patterns in addition to spatial structure. Apart from this temporal stacking, the pre-processing pipeline is identical to the 2D case. We use the same subset of simulations and time steps, apply the same normalization strategy, and store the resulting volumes in serialized format for efficient access during training and evaluation.

**Model Configuration.** The architecture of the 3D convolutional autoencoder mirrors that of the 2D model, with all two-dimensional operations replaced by their three-dimensional counterparts. Specifically, 3D convolutional and transposed convolutional layers are used to jointly learn spatial and temporal features from the input volumes. Kernel sizes, strides, and filter counts are chosen to match the 2D configuration as closely as possible, ensuring that differences in performance can be attributed primarily to the inclusion of temporal context rather than architectural complexity.

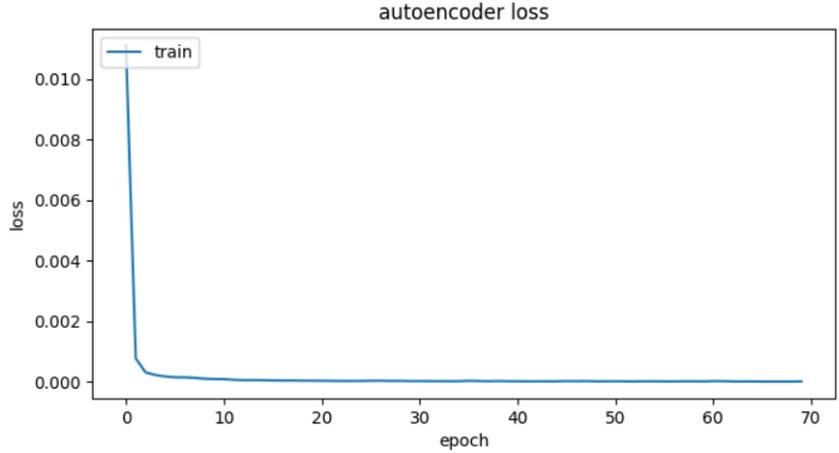

Figure 7: Validation loss in training the model

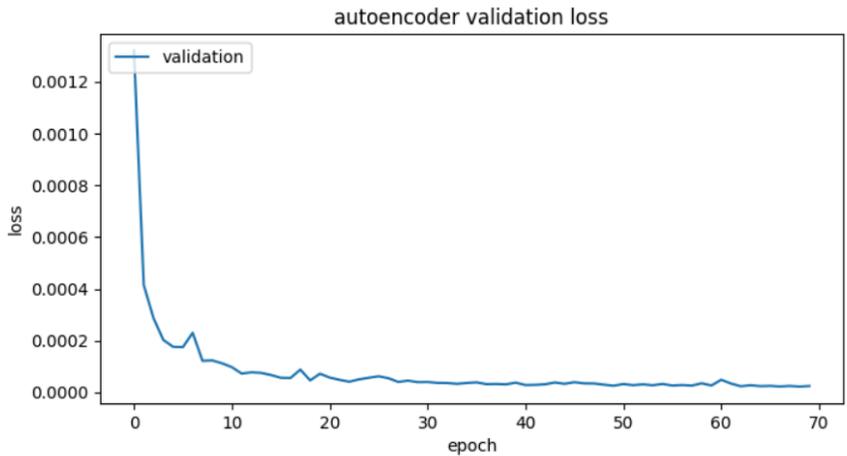

Figure 8: Validation loss in training the model

We visualize four times a stack of 3 images from volumes in the test set and the reconstruction thereof in Figure 9.

Details of the model construction are provided in Appendix C, and the full model summary generated by Keras is included in Appendix D.

**Training Procedure.** The 3D autoencoder is trained for 70 epochs using the same optimization strategy and reconstruction loss as the 2D model. Training and validation loss curves are shown in Figures 7 and 8, respectively. As in the 2D case, validation loss convergence is used as an indicator of sufficient training and to guard against overfitting.

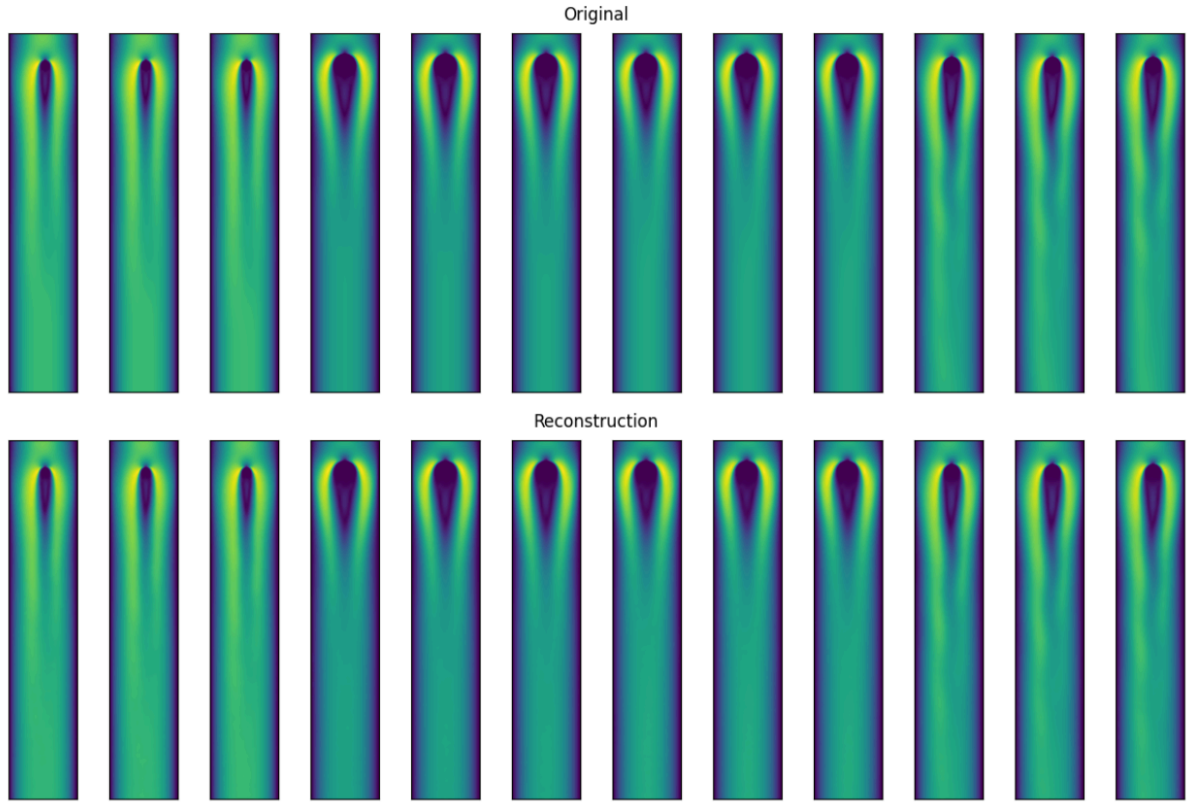

Figure 9: Reconstructions versus originals at 70 epochs

Testing and Anomaly Detection. To evaluate reconstruction performance, we compute the mean squared error for each volume in the test set. The distribution of reconstruction errors is shown in Figure 10. Similar to the 2D case, most samples exhibit low reconstruction errors, while a small number of volumes stand out with significantly higher values.

We apply the same anomaly threshold of $3 \times 10^{-4}$ to identify volumes with extreme reconstruction errors. These volumes are reconstructed using the trained 3D autoencoder, and representative examples are shown in Figure 11.

Qualitative inspection reveals that the detected anomalies largely correspond to dynamic irregularities in the flow evolution. Compared to the 2D model, which detects anomalies at the level of individual frames, the 3D autoencoder groups together temporally adjacent frames that exhibit consistent anomalous behavior. As a result, patterns that would appear as multiple, nearly identical detections in the 2D setting are consolidated into a single detection in the 3D case. This temporal aggregation reduces redundancy and enables a more coherent interpretation of anomalous motion patterns within the time series.

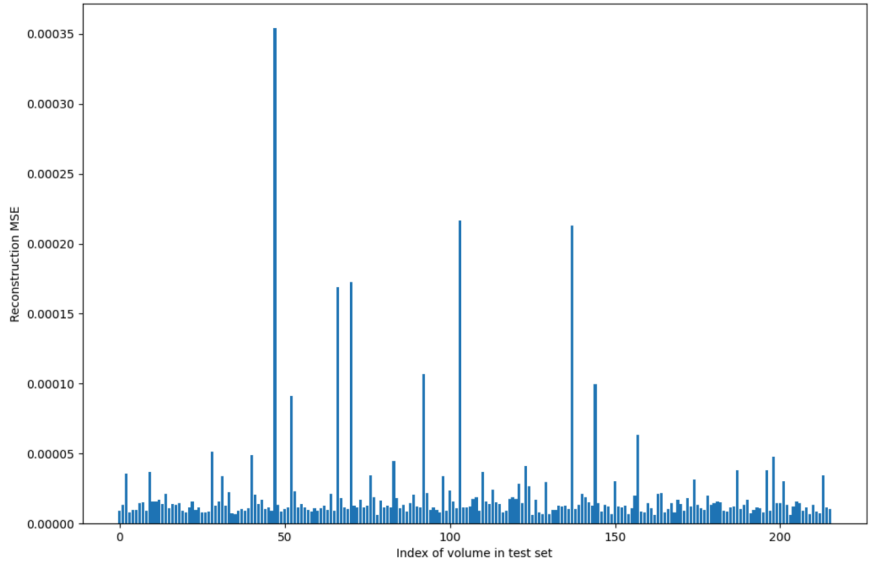

Figure 10: Reconstruction MSE for volumes in test set

### 3.3 3D Convolutional Autoencoder for Droplet3D Simulation

**Data Pre-processing.** In addition to the Kármán vortex street ensemble, we evaluate our approach on a fully three-dimensional, time-dependent flow simulation known as Droplet3D. This dataset is generated by the Institute of Aerospace Thermodynamics in Stuttgart and consists of volumetric simulations of droplet dynamics within a cubic domain. A visualization of selected time steps from the simulation is shown in Figure 12.

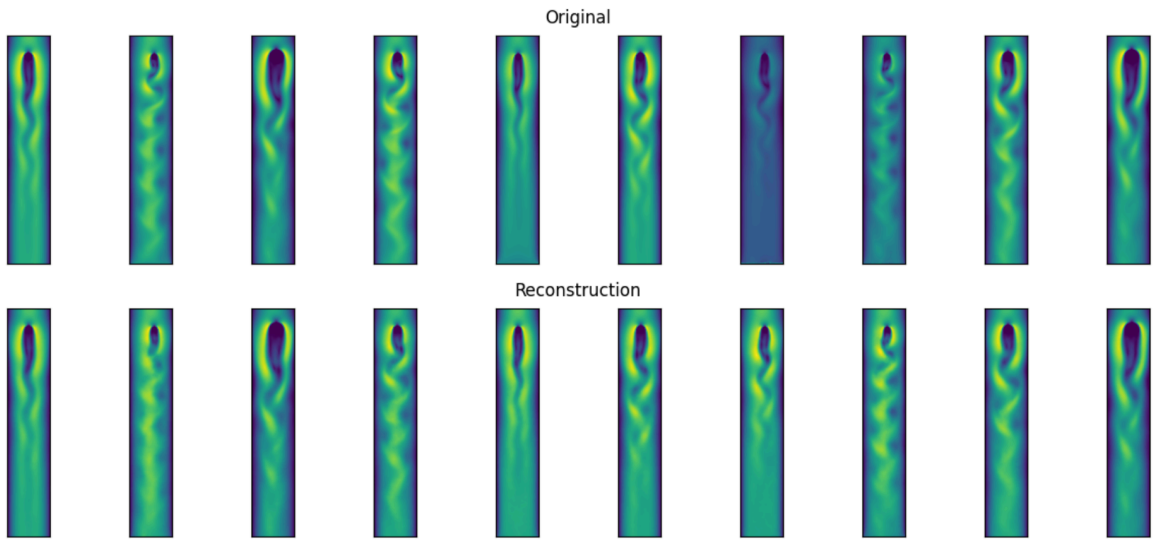

Figure 11: Reconstruction for volumes with high MSE in the test set

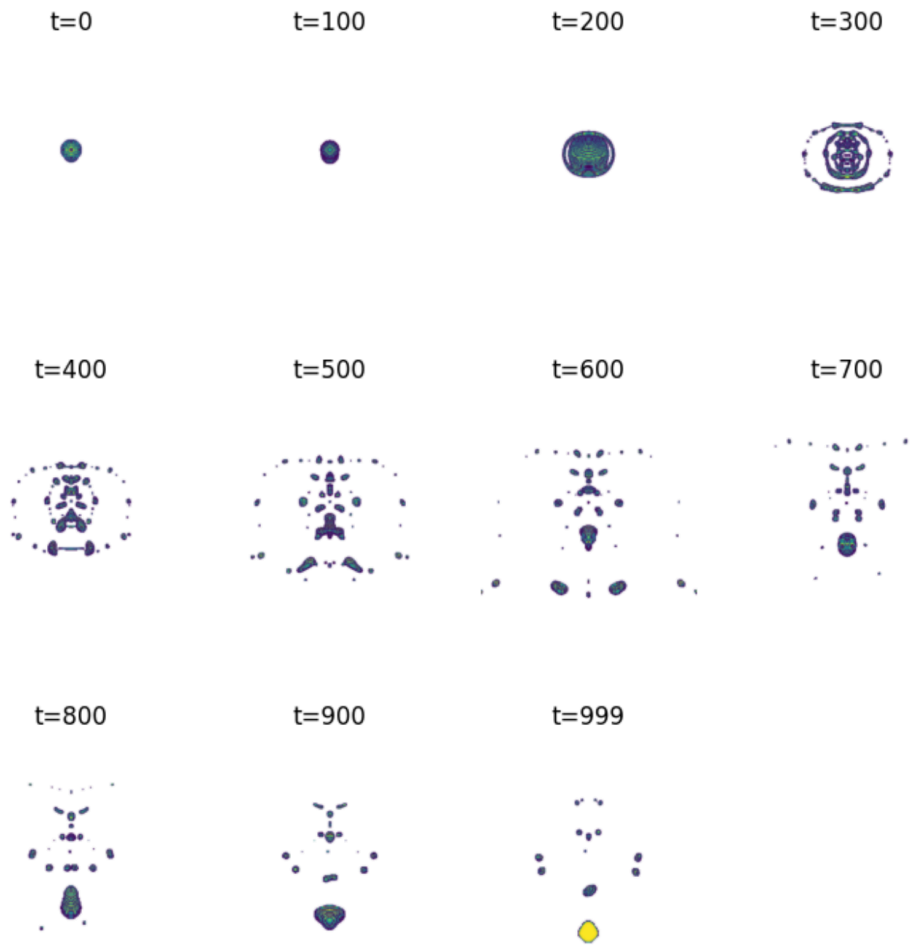

Figure 12: Visualization of a subset of the time steps in Droplet3D

The original data have a spatial resolution of 256 × 256 × 256. To make training computationally feasible, we downsample the volumes by a factor of two in each spatial dimension. We use the first 600 time steps of the total 1000-frame time series. Toward the end of the simulation, droplets increasingly disperse and leave the domain, resulting in near-empty volumes. At the final time steps, large portions of the volume contain no material, and droplets are only partially visible as they exit the simulation domain. To avoid biasing the model toward such trivial cases, we exclude the latter portion of the time series from training and evaluation.

In addition to spatial downsampling, the autoencoder further compresses the data to a latent representation of size 16 × 16 × 16. This two-stage reduction balances computational efficiency with the ability to capture relevant volumetric structures.

**Normalization and Training Challenges.** Applying the same normalization strategy used for the Kármán vortex street data—scaling values to the range [0, 1]—proved ineffective for the Droplet3D simulation. During training, the autoencoder consistently converged to a degenerate solution in which it reconstructed nearly empty volumes, regardless of the input. Further analysis revealed that this behavior is caused by the sparsity of the data: at most time steps, only a small fraction of the volume contains droplets, while the majority of voxels are empty.

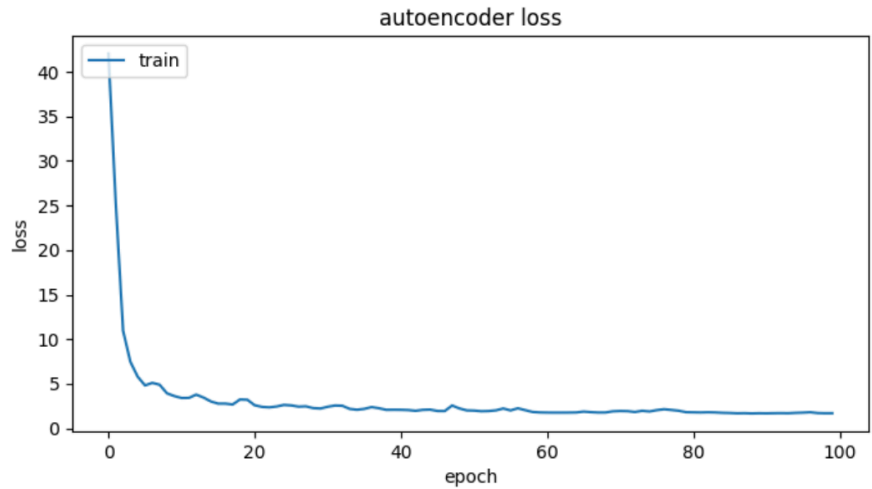

Figure 13: Validation loss in training the model

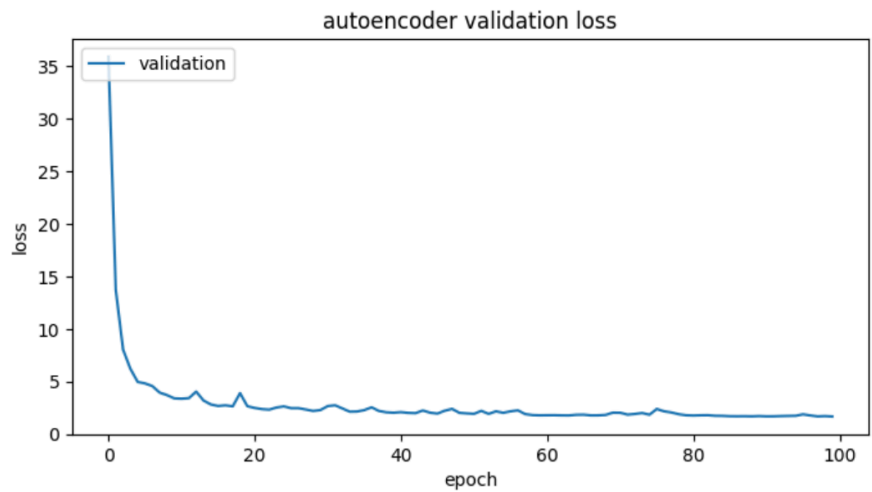

Figure 14: Validation loss in training the model

When normalized to [0, 1], reconstructing an empty volume results in a relatively low mean squared error, even when droplets are missing in the reconstruction. As a result, the optimization process becomes trapped in a local minimum where the model minimizes loss by ignoring sparse but physically important structures.

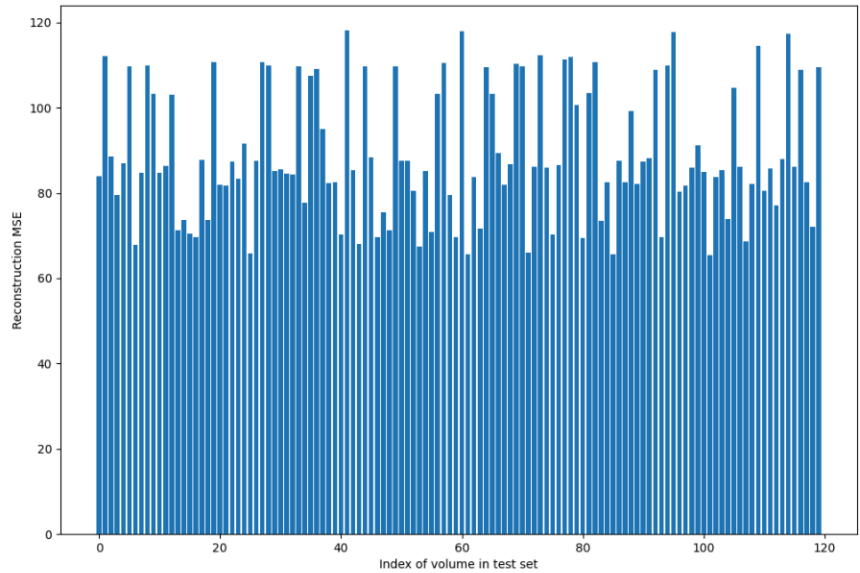

Figure 15: Reconstruction MSE for volumes in test set

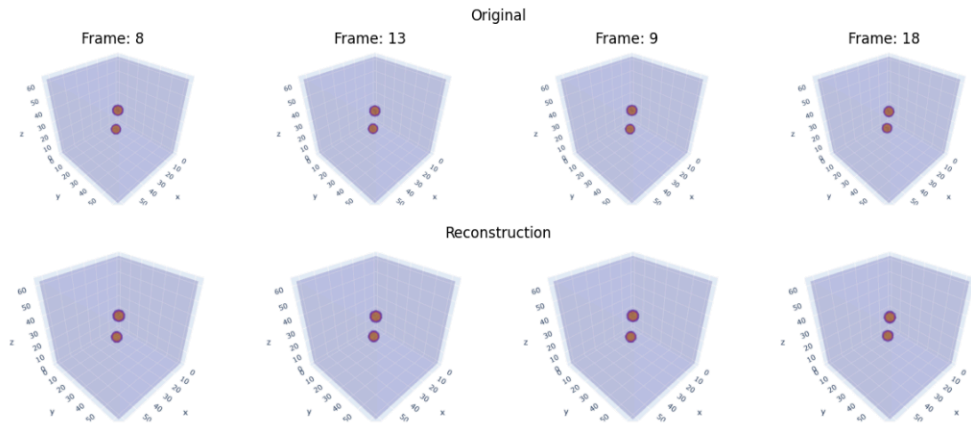

Figure 16: Reconstruction for volumes with high MSE in the test set

To address this issue, we retain the original data values in the byte range [0, 255] rather than applying min–max normalization. This increases the penalty associated with failing to reconstruct regions of high mass concentration and prevents the model from collapsing to trivial empty reconstructions. Using this representation, the autoencoder is able to learn meaningful volumetric features and produce reconstructions that preserve the presence and structure of droplets.

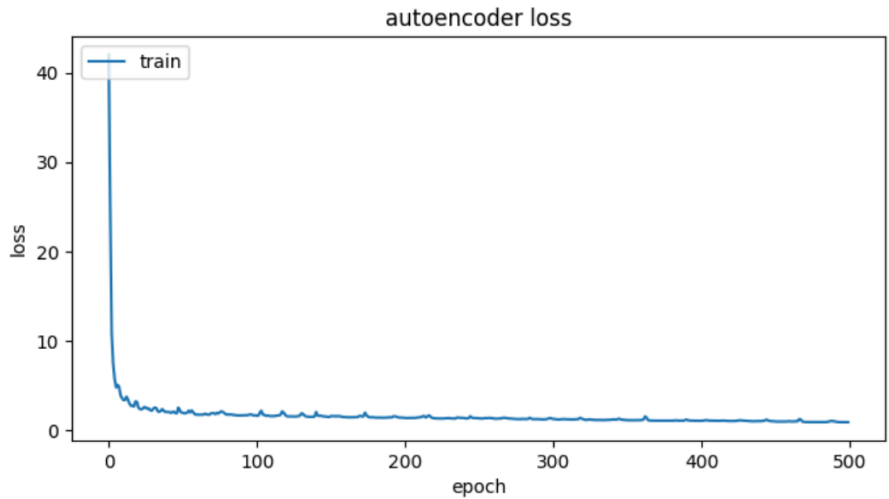

Figure 17: Validation loss in training the model

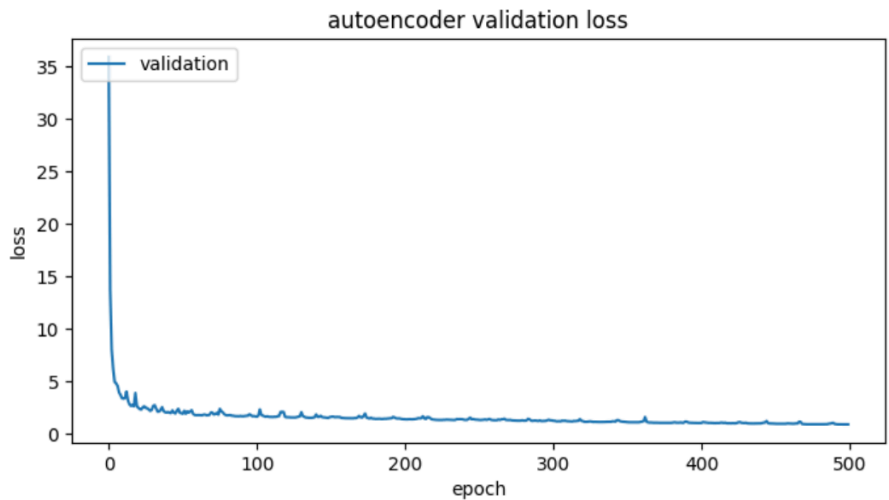

Figure 18: Validation loss in training the model

## 3.4 Model Configuration

The model configuration for the Droplet3D dataset largely follows the architecture used for the previous 3D convolutional autoencoder and is detailed in Appendix E, with the corresponding Keras model summary provided in Appendix F. As the input data are fully three-dimensional, convolutional kernels and strides are applied uniformly across all three spatial dimensions.

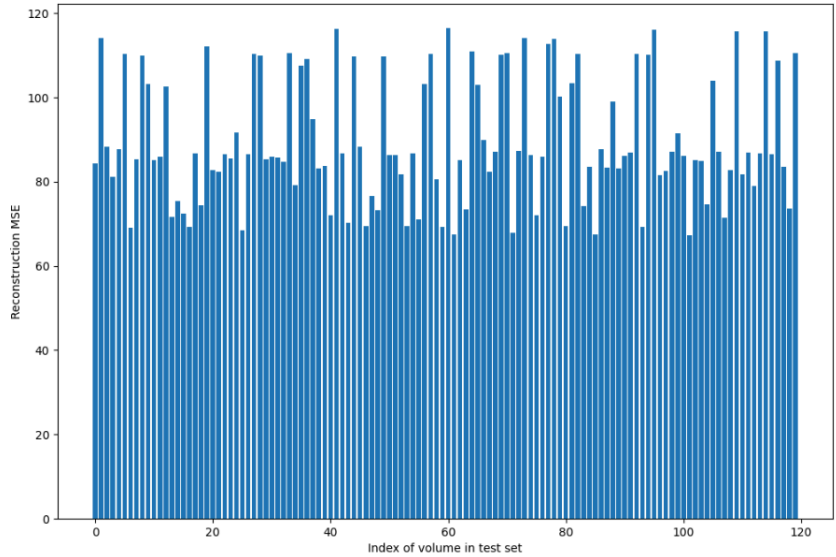

Figure 19: Reconstruction MSE for volumes in test set

In contrast to the stacked-frame approach used for the Kármán vortex street data, where the temporal dimension was treated differently, the Droplet3D simulation represents a true volumetric field. Consequently, downsampling is performed isotropically in all dimensions. To avoid excessive loss of information, we reduce the depth of the network and employ only two convolutional layers instead of four. Deeper architectures were found to introduce too much compression, resulting in poor reconstruction quality and loss of fine-scale droplet structures.

**Training.** The model is trained for 100 epochs using mean squared error as the reconstruction loss. The objective is to learn a faithful reconstruction of volumetric flow fields such that volumes exhibiting atypical structure or mass distribution yield elevated reconstruction errors. Training and validation loss curves are shown in Figure 13 and Figure 14, respectively. As in previous experiments, convergence of the validation loss indicates stable training behavior.

**Testing and Analysis.** After training, we compute the reconstruction mean squared error for all volumes in the test set. The resulting distribution of MSE values after 100 epochs is shown in Figure 15. A small number of volumes clearly stand out with substantially higher reconstruction errors.

To isolate these cases, we apply a threshold of 115, selecting the four volumes with the highest MSE values. The corresponding reconstructions are visualized in Figure 16, rendered from an oblique angle to better reveal the spatial structure of the spherical droplets present in the volume.

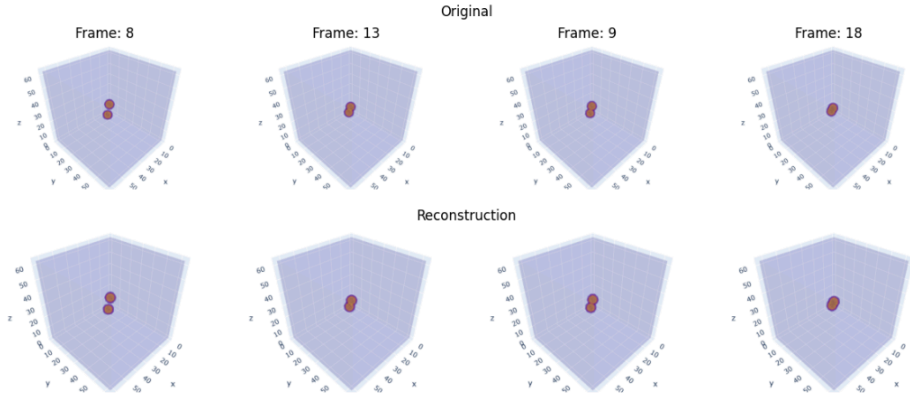

Figure 20: Reconstruction for volumes with high MSE in the test set

Interestingly, all detected anomalous volumes correspond to early time steps in the simulation, specifically time steps 8, 9, 13, and 18. This observation is notable, as anomalies could in principle occur anywhere within the selected temporal range of 0 to 600. Later time steps generally exhibit more complex droplet distributions, yet they are reconstructed more accurately by the model.

To investigate whether extended training improves discrimination, we continue training the model for up to 500 epochs. The corresponding training and validation loss curves are shown in Figures 17 and 18, and the resulting MSE distribution is shown in Figure 19. While the contrast between high and low reconstruction errors becomes slightly more pronounced, the set of detected anomalous volumes remains unchanged (Figure 20). Lowering the anomaly threshold results in the selection of additional volumes, but these are consistently confined to early time steps below approximately 120 and exhibit similar structural characteristics.

To better understand why these early volumes are difficult to reconstruct, we compare the volumetric state at time step 8 (Figure 21) with a representative later state at time step 300 (Figure 22). In early time steps, the data consist of two compact spherical droplets with a high concentration of mass localized in small regions of the domain. In contrast, later time steps exhibit more dispersed mass distributions, with droplets spread over a larger spatial area.

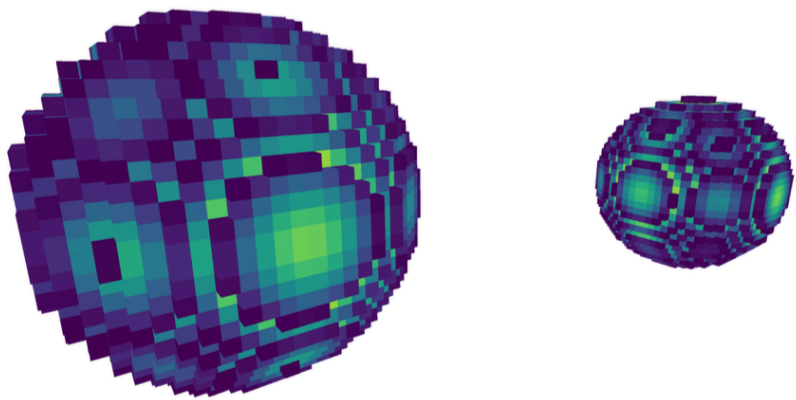

Figure 21: Visualization of the volume at time step 8

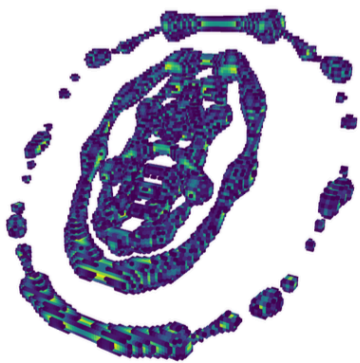

Figure 22: Visualization of the volume at time step 300

A close inspection of the droplet at time step 8 reveals a dense core containing near-maximum mass values, visualized in Figure 23. Such concentrated structures are inherently challenging for reconstruction-based models. Even a small spatial misalignment—on the order of a single voxel—between the reconstructed and original volume leads to large reconstruction errors due to the steep gradients between high- and low-value regions. Furthermore, localized high-magnitude structures occupy a relatively small portion of the latent representation compared to more dispersed configurations, making them harder to encode accurately. We therefore identify high mass concentration in compact regions as the primary factor contributing to elevated reconstruction errors in the Droplet3D dataset. This behavior highlights an important characteristic of reconstruction-based anomaly detection: anomalies are not solely defined by visual or structural complexity, but also by the spatial distribution and magnitude of values within the data.

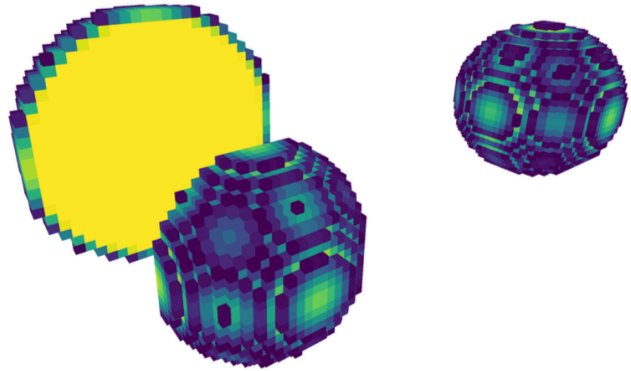

Figure 23: Visualization of a dissection of the volume at time step 8

## 4. Conclusion

In this work, we investigated reconstruction-based anomaly detection in ensemble and time-dependent simulation data using convolutional autoencoders. We compared a 2D convolutional autoencoder operating on individual simulation frames with a 3D variant that incorporates temporal context by stacking consecutive frames. While the 2D model effectively identifies spatial irregularities within individual images, the 3D model yields more meaningful detections by aggregating temporally related anomalies and reducing redundant detections of similar patterns within the same simulation.

We further applied a fully 3D convolutional autoencoder to volumetric Droplet3D simulation data. Our experiments demonstrate that reconstruction errors are strongly influenced by the spatial distribution of mass within the volume. In particular, volumes containing highly concentrated regions of large-magnitude values are more difficult to reconstruct accurately than volumes with more dispersed mass distributions, even when the latter appear visually more complex.

These findings underline the importance of incorporating spatial and temporal context in anomaly detection for scientific simulations and provide insight into the types of structures that reconstruction-based methods are most sensitive to. Future work may explore alternative loss functions or hybrid representations to improve robustness to localized high-magnitude features and further enhance interpretability of detected anomalies.